 \documentclass[tablecaption=bottom,pmlr]{jmlr}

\usepackage{booktabs}
\usepackage[load-configurations=version-1]{siunitx} 

\usepackage{multirow}

\theorembodyfont{\upshape}
\theoremheaderfont{\scshape}
\theorempostheader{:}
\theoremsep{\newline}

\jmlrvolume{148}
\jmlryear{2021}
\jmlrsubmitted{submission date}
\jmlrpublished{publication date}
\jmlrworkshop{NeurIPS 2020 Preregistration Workshop} 

\title{Evaluating Adversarial Robustness in Simulated Cerebellum}

  \author{\Name{Liu Yuezhang} \Email{yliuhl@ust.hk}\\
   \Name{Bo Li} \Email{bli@ust.hk}\\
   \Name{Qifeng Chen} \Email{cqf@ust.hk}\\
   \addr HKUST}

\begin{document}

\maketitle

\begin{abstract}
It is well known that artificial neural networks are vulnerable to adversarial examples, in which great efforts have been made to improve the robustness. However, such examples are usually imperceptible to humans, and thus their effect on biological neural circuits is largely unknown. This paper will investigate the adversarial robustness in a simulated cerebellum, a well-studied supervised learning system in computational neuroscience. Specifically, we propose to study three unique characteristics revealed in the cerebellum: (i) network width; (ii) long-term depression on the parallel fiber-Purkinje cell synapses; (iii) sparse connectivity in the granule layer, and hypothesize that they will be beneficial for improving robustness. To the best of our knowledge, this is the first attempt to examine the adversarial robustness in simulated cerebellum models. 

The results are negative in the experimental phase---no significant improvements in robustness are discovered from the proposed three mechanisms. Consequently, the cerebellum is expected to be vulnerable to adversarial examples as the deep neural networks under batch training. Neuroscientists are encouraged to fool the biological system in experiments with adversarial attacks. 
\end{abstract}


\section{Introduction}
\label{sec:intro}

The recent renaissance in deep learning has achieved great success in various challenging supervised learning tasks, which shows the enormous capacity of neural networks. Intriguingly, adversarial examples (i.e., intentional designed small perturbations which could fool machine learning systems) were discovered \cite{szegedy2013intriguing, goodfellow2014explaining}, providing the robustness as an additional criterion. Such adversarial examples are usually undetectable to the human visual system but destructive to artificial neural networks, known as the \textbf{robustness gap} between biological and artificial learning systems \cite{carlini2019evaluating}.

Significant efforts have been made in adversarial defense from the engineering perspective \cite{madry2017towards, dhillon2018stochastic, balunovic2020adversarial}, while on the other side, robustness in the biological system is not fairly understood. \citet{nayebi2017biologically} proposed a defense inspired by saturated neurons, but was later breached with numerically stabilized gradients \cite{brendel2017comment}. \citet{li2019learning} discovered that regularization fitting neuron responses had a positive effect on robustness, while other studies  \cite{elsayed2018adversarial, zhou2019humans} suggested that adversarial examples could also fool time-limited human based on the behavior experiments. In all, concrete results of adversarial robustness \textbf{on the neural circuits level} is lacking investigation.

Under ideal conditions, we should directly study the robustness of a visual system. However, as ``an active field of research'' \cite{dapello2020simulating}, it seems to be infeasible to evaluate the robustness of a simulated visual model for the moment---a supervised learning system in nature with detailed understandings ought to be a better candidate for our purpose at the current stage. 

The cerebellum is one of the most studied brain areas in computational neuroscience. The standard model, known as the Marr-Albus theory \cite{marr_cerebellum_1969, albus_cerebellum_1971}, proposed the cerebellum as a perceptron network for movements learning. Although various amendments have been made to the original idea, the supervised paradigm of cerebellar learning is well established \cite{apps2009cerebellar, chaumont2013clusters, raymond2018computational}. Therefore, as a supervised learning system in nature, it would be intriguing to ask: does the adversarial examples problem also exist in the cerebellum?

We hypothesize that the unique architectures in cerebellar circuits\footnote{We use the rat cerebellum as an exemplar through our study.} will enhance the robustness, and thus, such adversarial vulnerabilities in artificial neural networks are not expected to exist in the cerebellum model. To test our idea, we propose to simulate the cerebellum and evaluate its robustness against adversarial attacks. Admittedly, addressing such a problem is in some sense betting on both sides: if the results are affirmative, the mechanisms improving robustness will be inspiring for adversarial defense; otherwise, adversarial examples are expected to be discovered in vivo by neuroscientists. Given the significance of both positive and negative results, our work is especially suitable for a pre-register study.

\section{Related Work}
\label{sec:review}

\paragraph{Computational models of cerebellum}The long trend for modeling the cerebellum in computational neuroscience was initiated by \citet{marr_cerebellum_1969}, in which he proposed the cerebellum as a perceptron model under Hebbian learning rule. \citet{albus_cerebellum_1971} presented a similar model independently with an emphasis on the depression activity. The proposed plasticity was experimentally verified by \citet{ito1982climbing}, and thereby, the Marr-Albus-Ito theory of cerebellum was established. Some of the following studies focused on the connectivity in the cerebellum. Namely, \citet{cayco2017sparse} suggested that the sparse connectivity was essential for pattern decorrelation by simulation, and \citet{litwin2017optimal} further derived the optimal degrees of connectivity in cerebellum-like structures from a theoretical perspective. Other studies attempted to rebuild the cerebellum and test its capabilities in simulation. \citet{hausknecht2016machine} evaluated machine learning capabilities of a simulated cerebellum, addressing that such a model can solve simple supervised learning benchmarks such as MNIST. Yamazaki's group also has a long history in building large scale cerebellar models with either GPU~\cite{yamazaki2013realtime} or supercomputer~\cite{yamaura2020simulation}, representing the state-of-the-art in computational modeling of the cerebellum. However, as a newly developed topic from the machine learning community, the adversarial robustness of such models has not yet been examined.

\paragraph{Adversarial defense}Much effort has been devoted to adversarial defensive techniques. Adversarial training \cite{goodfellow2014explaining} was proposed as an efficient framework to improve the robustness. \citet{madry2017towards} further equipped the framework with a more advanced attack. Gradient masking effect was discovered \cite{papernot2016towards, tramer2018ensemble, athalye2018obfuscated}, which showed that a bunch of methods except adversarial training indeed leveraged such effect and provided a false sense of security. Randomization methods are the second type of defensive techniques acquiring comparable success against black-box attacks. Randomness was introduced to input transformation \cite{xie2018mitigating}, layer activation \cite{liu2018towards}, or feature pruning \cite{dhillon2018stochastic}. Unfortunately, such methods are facing challenges with white-box attacks \cite{athalye2018synthesizing}. Provable defense \cite{wong2018scaling} is another trend aiming at proposing algorithms with guaranteed safety, but usually with the cost of significantly reducing the training accuracy. A recent work \cite{balunovic2020adversarial} attempted to bridge this gap in provable defense, but also benefits from the adversarial training. In all, adversarial training is still the most successful defensive technique developed so far. However, we challenge if such a framework is biologically plausible---no evidence supports that the biological neural circuits process information augmented with adversarial signals repeatedly to enhance its robustness. In pursuit of a machine learning system with human-level robustness, intriguing defensive mechanisms other than adversarial training should be expected in the brain.

\section{Methodology}
\label{sec:method}
\subsection{Anatomy of Cerebellum}
\begin{figure}[htbp]
	\floatconts
	{fig:subfigex}
	{\caption{\textbf{The schematic diagrams of cerebellum.} (a) Anatomical schematic. MF: mossy fiber, GC: granule cell, PC: Purkinje cell, CF: climbing fiber, LTD: long-term depression. (b) Computational model.  Please note the corresponding relationship between the anatomical and computational model.}}
	{%
		\subfigure[Anatomical schematic.]{\label{fig:anatomy}%
			\includegraphics[width=0.45\linewidth]{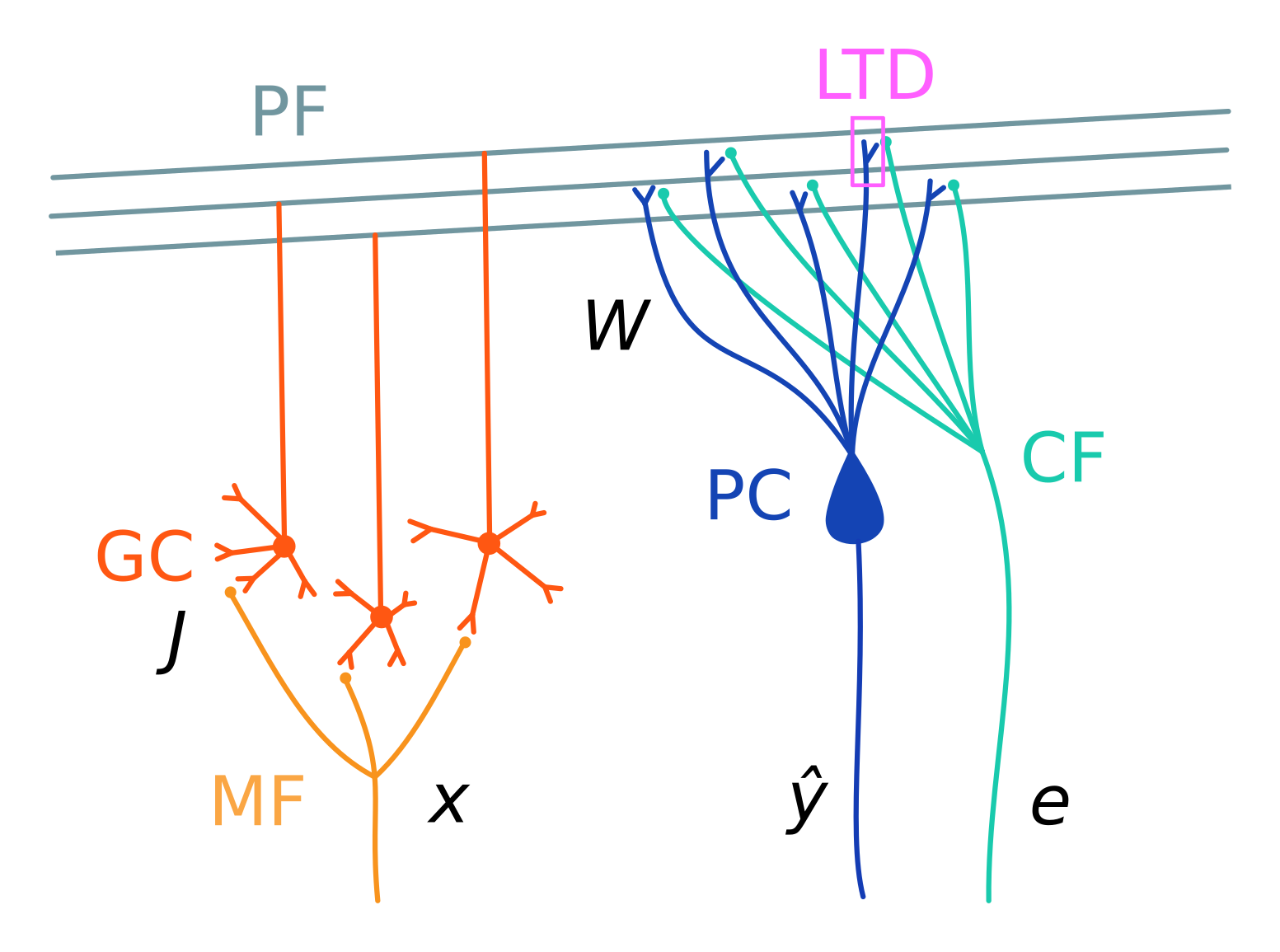}}%
		\qquad
		\subfigure[Computational model.]{\label{fig:model}%
			\includegraphics[width=0.45\linewidth]{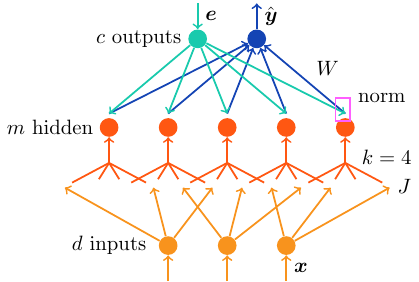}}
	}
\end{figure}

We briefly summarize several anatomical pieces of evidence to justify our model of cerebellum. Connectivity in the cerebellar cortex is substantially feed-forward: information inputs through the mossy fiber (MF, input neuron) to granule cell (GC, hidden neuron), and subsequently passes through parallel fiber (PF) to Purkinje cell (PC, readout neuron). Another input comes from climbing fiber (CF) to PF-PC synapse. The schematic diagram of cerebellum is shown in Fig~\ref{fig:anatomy}. In quantity, corresponding to one single PC output, there are approximately $d=7,000$ MFs, $m=200,000$ GCs, and one PC in the rat brain \cite{marr_cerebellum_1969}. By connectivity, the MF-GC layer is sparsely connected---one GC receives $k=4$ MF inputs on average \cite{litwin2017optimal}, while the PF-PC layer is densely connected---up to 200,000 GCs are connected with one single PC output.

The special one-to-one relationship between the CF and PC motivated Marr to predict that the PF-PC synapses are facilitated by the presynaptic and CF (postsynaptic) activities with Hebbian learning rule, and no other synapses are modifiable. Later studies showed that the cerebellar nuclei (CN, the output from PC) and inferior olive (IO, the input to CF) form a close loop to generate instructive signals, and CFs are carrying error signals \cite{apps2009cerebellar, raymond2018computational}, which essentially established the error propagation mechanism through a single layer in the cerebellar cortex. Long-term depression (LTD) \cite{ito1982long, ito1989long, hirano2013long} was also discovered on the PF-PC synapse---reduction in the efficacy occurs in excitatory synapses and lasting for hours after receiving a long series of stimulus.

Note that based on the listed facts, backpropagation (BP) through multiple layers is unlikely to be applicable in the cerebellum. The mechanism of error propagation is well established, which relies on the special one-to-one relationship of PC and FC. Since the feedback fibers are not discovered beyond the Purkinje layer, the cerebellum may not propagate the error signals to the granule layer or further, unless different mechanisms \cite{guerguiev2017towards} have been evolved for the same function of error propagation in the cerebellum.

The absence of multi-layer BP in the cerebellum does not effect applying gradient-based adversarial attacks, as our model could be regard as a combination of fixed random projection and a linear readout trained with (single-layer) BP, i.e., is still end-to-end differentiable (see Sec.~\ref{sec:model}). Our preliminary experiments also confirm that such a setting does not provide additional benefits on robustness comparing to the conventional BP.

\subsection{Model Setup}
\label{sec:model}
Following Marr's theory, the cerebellum could be modeled as a simple two-layer rate neural network, as shown in Fig \ref{fig:model}. Consider the inputs $\boldsymbol{x}\in\mathbb{R}^d$ on MFs, the granule layer responses are given by
\begin{equation}
\boldsymbol{h}=\sigma(J^T\boldsymbol{x}),
\end{equation}
where $J\in \mathbb{R}^{d\times m}$ is the granule synaptic weights matrix which projects inputs to $m$ dimensional space, and $\sigma(\cdot)$ is the ReLU activation function. Note that $J$ is a sparse matrix---each row contains only $k$ non-zero entries. The indices of the non-zero entries are randomly selected (randomly connected, RC). The random connection setting is consider to be acceptable both experimentally \cite{gruntman2013integration} and theoretically \cite{litwin2017optimal}. The non-zero weights are initialized by Xavier initialization, thus $J_{ij}\sim\frac{1}{\sqrt{k}}\mathcal{N}(0, 1)$, and fixed after initialization.

The Purkinje layer outputs are given by
\begin{equation}
\hat{\boldsymbol{y}}=W^T\boldsymbol{z},\quad\boldsymbol{z}=\boldsymbol{h}-\boldsymbol{\mu},
\end{equation}
where $W\in\mathbb{R}^{m\times c}$ is the Purkinje synaptic weights matrix which maps the responses to $c$ classes and $\boldsymbol{\mu}$ is the normalization term. Since LTD is discovered in PF-PC synapses, we propose to model it as the normalization technique in machine learning. For the sake of biological plausibility, we consider the simple centralization with moving average, thus $\boldsymbol{\mu}_t=\beta\boldsymbol{\mu}_{t-1} + (1-\beta)\boldsymbol{h}_t$.

Here we wish to remark on the reasons we model the LTD as the normalization mechanism. Although the specific functional role of LTD is still unclear, we argue that it shares the same spirit as the ``activity-dependent reduction in synapses efficiency through a long time'' with normalization techniques in ML, which are usually found critical to reflect biological phenomena~\cite{coates2011analysis}. Our preliminary results also suggest such normalization may be useful to enhance robustness.

Now we describe the learning rule in PC. For some training instance $(\boldsymbol{x},\boldsymbol{y})$, consider the MSE loss
\begin{equation}
l=\frac{1}{2}\|\hat{\boldsymbol{y}}-\boldsymbol{y}\|_2^2=\frac{1}{2}\sum_{j=1}^c(\hat{y_j}-y_j)^2.
\end{equation}
By gradient descent, the updates of Purkinje synaptic weights
\begin{equation}
\Delta W \propto \frac{\partial l}{\partial W}=\boldsymbol{e}^T\boldsymbol{z},
\end{equation}
where $\boldsymbol{e}=\hat{\boldsymbol{y}}-\boldsymbol{y}$ is the error vector transmitted by the CF. Such subtraction error calculation could be supported by the inhibitory connection in the CN-IO loop. Note that the update rule is consistent with Hebbian learning---as synaptic updates are proportional to the product of pre-($\boldsymbol{z}$) and post-($\boldsymbol{e}$) synaptic activities.

\section{Experimental Protocol}
\label{sec:experiment}
We now present the experimental protocol on evaluating the adversarial robustness of our cerebellum model. We will focus on three primary characteristics revealed in anatomy as discussed above:
\begin{itemize}
	\item \textbf{Network width.} Why does the cerebellum develop the shallow and wide architecture for supervised learning? Will the shallow and wide structure provide extra benefits compared with deep networks?
	\item \textbf{Long-term depression.} What are the potential benefits of the LTD effect in PF-PC synapses on robustness?
	\item \textbf{Sparse connectivity.} Will the extra sparse connectivity ($k=4$ vs. $m=200,000$) in the granule layer provide advantages on robustness compared with fully-connected or other configuration of $k$?
\end{itemize}

All the code, pre-trained models, and scripts necessary to reproduce our results will be made publicly available in the results paper.

\paragraph{Robustness measurement}The robustness is defined as the stability against input perturbations. A learning system sensitive to small perturbations will face great challenges in real-world environments with noisy nature. Although we could in general measure robustness with arbitrary disturbances \cite{yang2019evolving}, adversarial examples provide a more rigorous evaluation estimating the worst-case robustness \cite{carlini2019evaluating}. Therefore, we propose to study the adversarial robustness in the proposal, and use the perturbed accuracy under adversarial attacks as the robustness metric.

\paragraph{Primary datasets and training setup}
To keep the results comparable with machine learning researches, we will conduct our evaluations on the MNIST and CIFAR-10 datasets. Prior studies \cite{hausknecht2016machine, illing2019biologically} have already shown that the plain cerebellum-like architectures are capable of solving the MNIST dataset, therefore no additional modifications are needed for our cerebellum model on MNIST. We will use MNIST as the primary benchmark for the experiments in network width, LTD, and sparse connectivity. For the final evaluation of cerebellum-like models and ablation studies, we will include results on both MNIST and CIFAR-10.

However, since the cerebellum does not involve any receptive field (convolution) structures, it could be challenging to apply the model to CIFAR-10 without modification. Here we plan to deploy a visual feature encoder to resolve the problem. Specifically, we will first train a standard VGG-11 model (8 convolution and 3 fully-connected layers) as the baseline. We will apply standard data augmenting, such as random cropping and horizontal flipping during baseline training. After acquiring a desirable visual model of CIFAR-10, we will utilize the 8 convolution layers (discarding FC layers) as the visual feature encoder, mapping the CIFAR-10 dataset to a representation space ($d=4096$). We will further train our cerebellum model on top of the visual feature encoder, keep the convolution layers fixed while training the cerebellum. Attacking will also be applied to the same model with the visual encoder and the cerebellum backend to keep the perturbation level ($\epsilon$) consistent in the original image space.

\paragraph{Secondary training tasks}
Despite the MNIST and CIFAR-10, we could also conduct more biologically realistic tasks for modeling cerebellar learning, such as eyelid conditioning~\cite{medina2000computer} or pole balancing through reinforcement learning~\cite{hausknecht2016machine}. However, since such tasks are not widely applicable in adversarial learning research, their significance for evaluating adversarial robustness is relatively limited (for instance, it will be hard to define proper perturbation level and make straightforward comparisons across models). Furthermore, those tasks are found less challenging than MNIST~\cite{hausknecht2016machine}. We here wish to justify our motivation for performing ``visual recognition'' tasks with the cerebellum model--the consideration is more from adversarial robustness evaluation than the biological plausibility. We will append the experiments on eyelid conditioning and pole balancing as secondary results if time permits.

\paragraph{Threat models}
We will consider two white-box attacks under $l_\infty$ norms as our threat models: the simple FGSM \cite{goodfellow2014explaining} attack, as well as more advanced PGD \cite{madry2017towards} attack. For MNIST, we consider $L_{\infty}$ perturbation $\epsilon=0.1,\,0.3$; for CIFAR-10, we consider $L_{\infty}$ perturbation $\epsilon=2/255,\,8/255$. The attack settings are standard practices \cite{madry2017towards, balunovic2020adversarial} and also recommended by \citet{carlini2019evaluating}.

\paragraph{Network width}
The first aspect we propose to study is the effect of network width on adversarial robustness. Following the previous setup, we consider the fully-connected (FC) model in the granule layer with a different number of GCs ($m$ ranges from 1000 up to 50,000). We are expected to observe that the network robustness increases along with its width, as reported by \citet{madry2017towards}. However, the significance of such an effect is still unclear. Is the wide and shallow structure a decisive factor to improve adversarial robustness in the cerebellum? Or, is it simply a biological drawback as the cerebellum fails to evolve the sophisticated backpropagation algorithm?

\paragraph{Long-term depression}
To evaluate the effect of LTD, we will repeat the experiments for evaluating the effect of network width. The only difference is that we will apply normalization techniques (moving average) in the PC-FC synapses. We wish to study whether the LTD mechanism will improve the robustness in the cerebellum model.

\paragraph{Sparse connectivity}
We will next investigate sparse connectivity in the granule layer. We propose to consider different levels of sparsity, namely $k = 1, 2, 4, 10, 50, 200, 784$ (FC). For a fair comparison, we will control the total number of synapses in the models, as in \citet{litwin2017optimal}. Since each GC receives $k$ MF inputs, and corresponds to 1 PC and 1 CF synapse for each output ($c$ classes in total), the number of synapses is given by $m(k+2c)$. For the cerebellum model, $k=4$, $m=200,000$, and $c=10$ is decided by anatomy and datasets, thus the number of synapses is $4.8M$, and the number of GCs is given by $m=4.8M/(k+2c)$. In this experiment, we wish to understand if sparse connections in the granule layer will bring advantages to robustness in the cerebellum.

\paragraph{Cerebellum models and ablation studies}
By adding the LTD and sparse connectivity or not, we could eventually evaluate the robustness across four cerebellum-like models: from the base model which is fully connected in the granule layer without LTD ($k=d=784,\,m=5970$ on MNIST, $k=d=4096,\,m=1166$ on CIFAR-10), to the cerebellum model which is sparsely connected with LTD ($k=4,\,m=200,000$). We will also examine whether the parameters decided by the natural cerebellum ($k=4$) is (near) optimal against adversarial attacks.

\paragraph{Hyperparameters}
Hyperparameters for training and attacking the cerebellum model are listed in Table~\ref{tab:init-hyper-cereb}. The initially proposed values are based on previous studies \cite{illing2019biologically, madry2017towards, balunovic2020adversarial} and our preliminary experiments. We will tune the hyperparameters during attacking to ensure that our evaluation does not suffer from gradient masking (see discussion below).

\begin{table}[htbp]
	\caption{Initially proposed hyperparameters for evaluating the cerebellum model.}
	\label{tab:init-hyper-cereb}
	\medskip
	\centering
	\begin{tabular}{p{0.25\linewidth}c}	
		\toprule[1.5pt]
		\textbf{Hyperparameter} & \textbf{Value} \\
		\midrule[1pt]  
		Training batch size & 10 \\
		Epoch            & 10 \\
		Learning rate    & $0.5/m$ \\
		Optimizer        & RMSprop \\
		Optimizer decay  & 0.99 \\
		LTD $\beta$      & 0.99 \\
		Seed             & \{0, 123\} \\
		\midrule[1pt]
		Attacking batch size & 10 \\
		PGD steps        & 40 \\
		PGD step size    & 0.01 \\
		\bottomrule[1.5pt]
	\end{tabular}
\end{table}

\paragraph{Gradient masking and robustness evaluation}
Gradient masking is the major concern in applying gradient-based iterative methods, which will provide a false sense of robustness if not examined properly \cite{carlini2019evaluating}. The FGSM attack will be used as a sanity test for the PGD attack, as iterative attacks should always perform better than single-step methods. Gradient distribution plots in attacking will be utilized to avoid gradient masking. We will carefully tune the attacking hyperparameters and perform a doubling iterations test to certify that the PGD attack converges (i.e., the fooling rate does not further increase with respect to the number of iterations). We will also conduct an unbounded attack to ensure that our attacking methods have been applied properly.

\paragraph{Random seeds and transferability analysis}
Since our models involve randomness (especially in the RC granule layer), we propose to report results with different random seeds. We will also perform a transferability attack and analysis across different initialized models \cite{papernot2016transferability, madry2017towards}. Better robustness against transferable adversarial examples is also expected in the cerebellum model.

\section{Results}
\label{sec:result}
We next report the experimental results based on the proposed protocol. The performance of the deep learning baseline is marked in gray dashed lines in figures. Experiments are repeated with $2$ random seeds. Hyperparameters and original data are listed in the appendix. The codebase, pre-trained models, and training scripts necessary to reproduce our results are available at \url{https://github.com/liuyuezhang/cerebellum}.

\begin{figure}[htbp]
	\floatconts
	{fig:subfigex}
	{\caption{\textbf{The effect of network width, LTD and sparse connectivity on adversarial robustness.} The gray dashed represents the performance of the deep learning baseline. (a) Network width slightly improves robustness against low perturbation iterative attack. No advantages are revealed by taking the shallow and wide architecture comparing with the deep model. (b) Similarly, LTD also only has a limited improvement on robustness on low perturbation with different widths (here $m=2000$ and $200,000$ are presented). (c) The effect of sparse connectivity on adversarial robustness. The x-axis represents the sparsity ($k$, ranges from 1 to 784). Sparse connections in the granule layer show advantages on robustness to full connections, but again the effect is relatively subtle.}\label{fig:result-mnist}}
	{%
		\subfigure[Network width.]{\label{fig:width}%
		\begin{minipage}[t]{\linewidth}
			\includegraphics[width=0.43\linewidth]{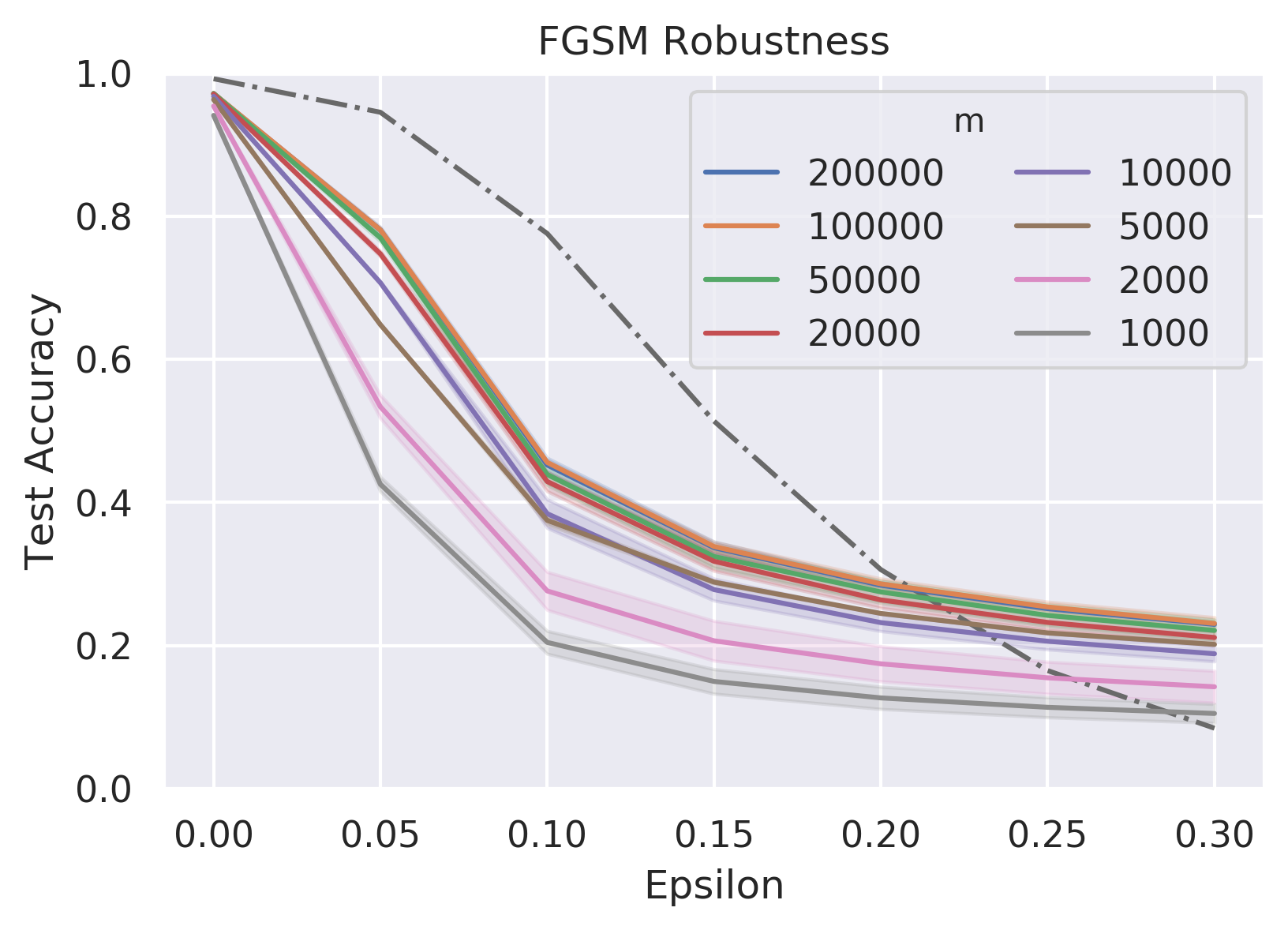}
			\hspace{2.5em}
			\includegraphics[width=0.43\linewidth]{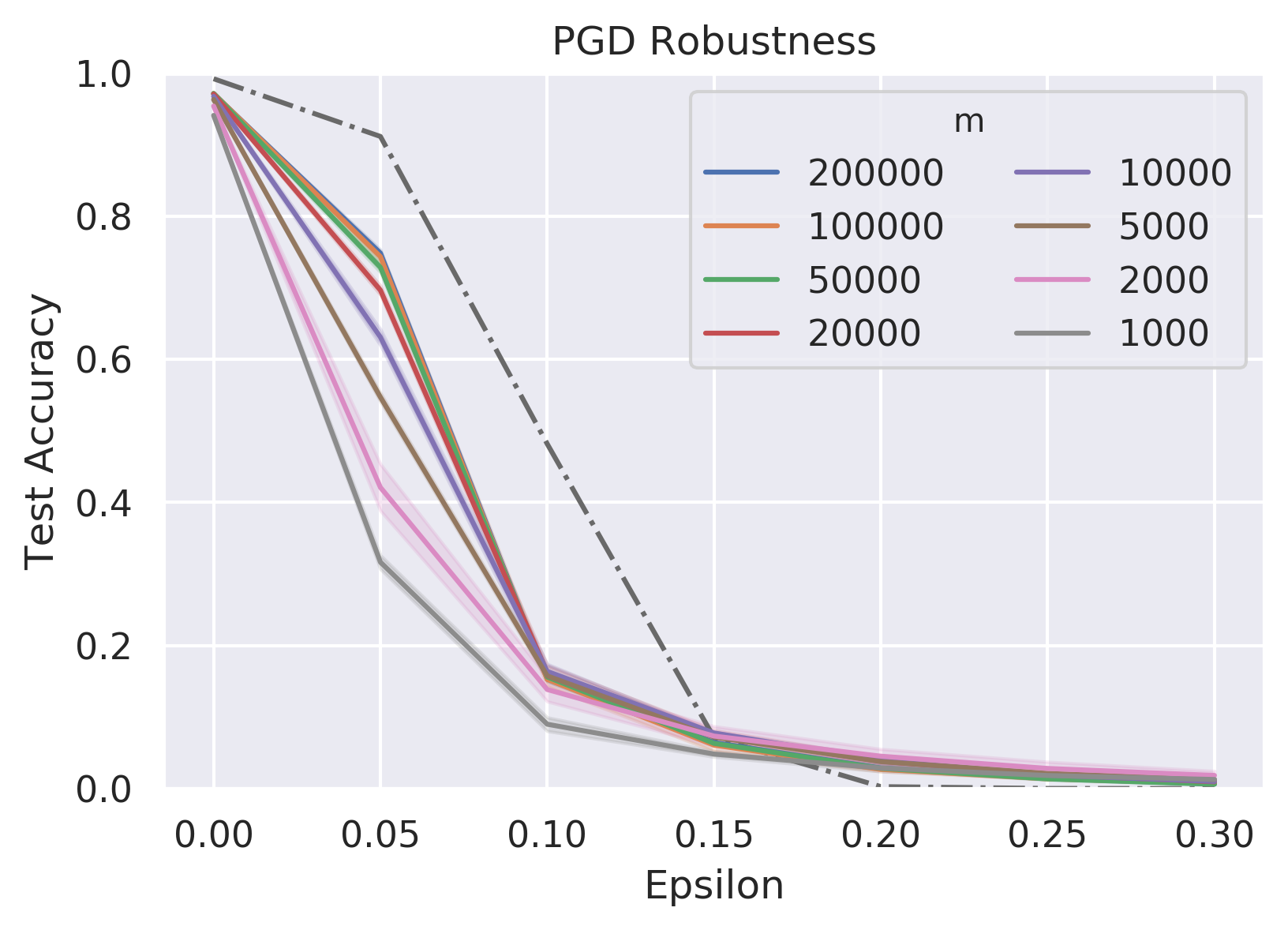}
		\end{minipage}} \\
		\subfigure[LTD.]{\label{fig:ltd}%
		\begin{minipage}[t]{\linewidth}
			\includegraphics[width=0.43\linewidth]{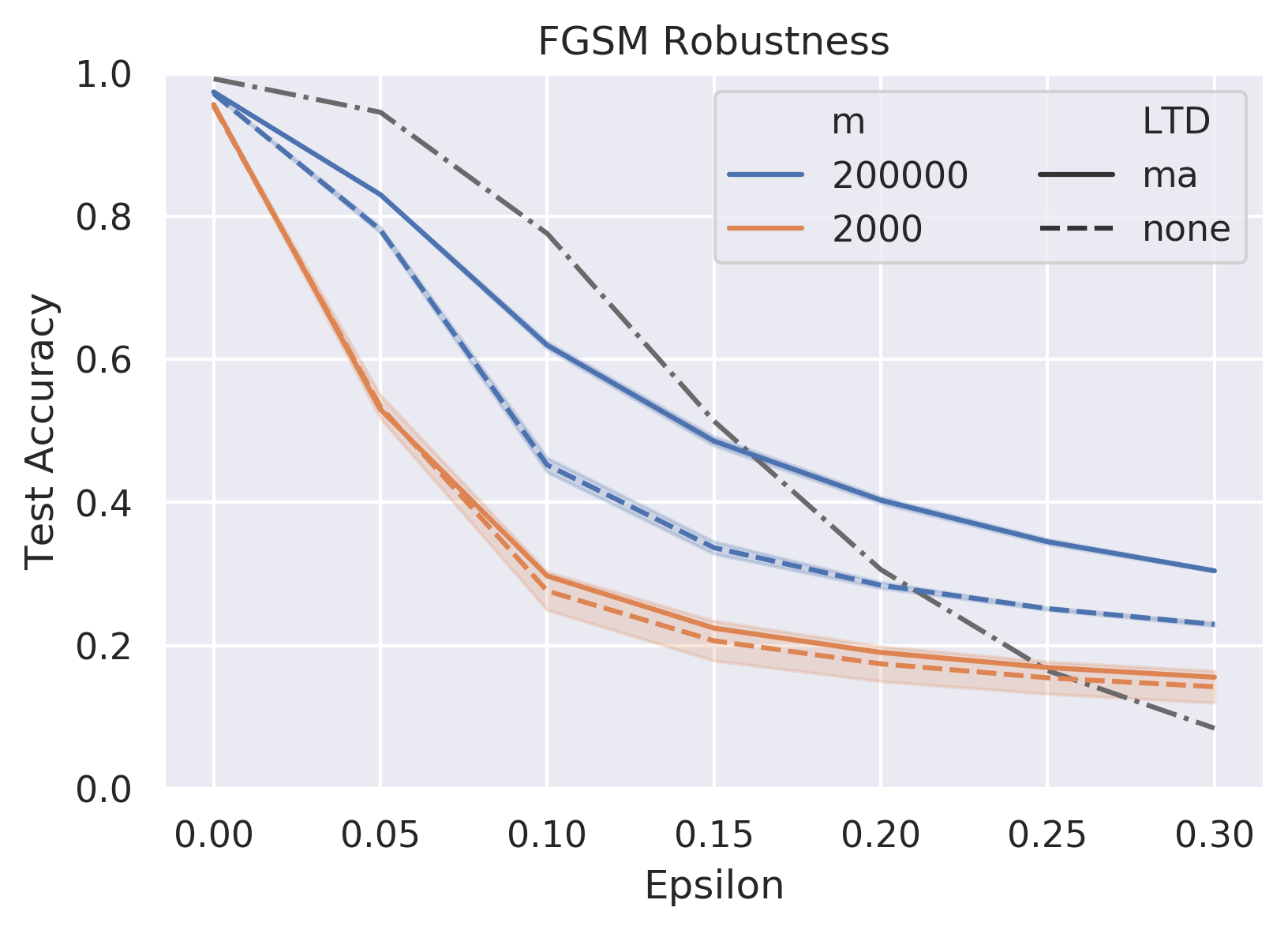}
			\hspace{2.5em}
			\includegraphics[width=0.43\linewidth]{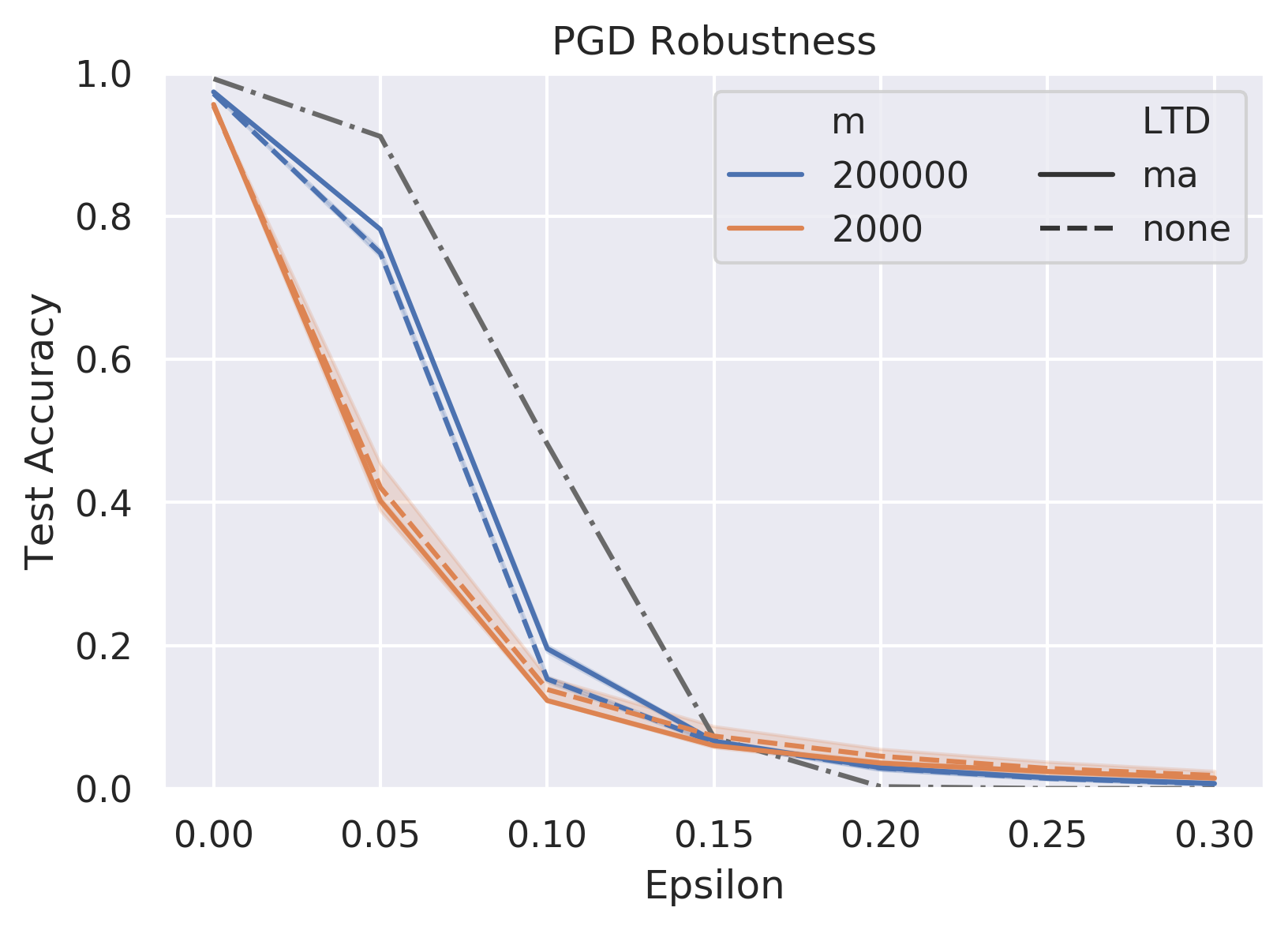}
		\end{minipage}} \\
		\subfigure[Sparse connectivity.]{\label{fig:sparse}%
		\begin{minipage}[t]{\linewidth}
			\includegraphics[width=0.5\linewidth]{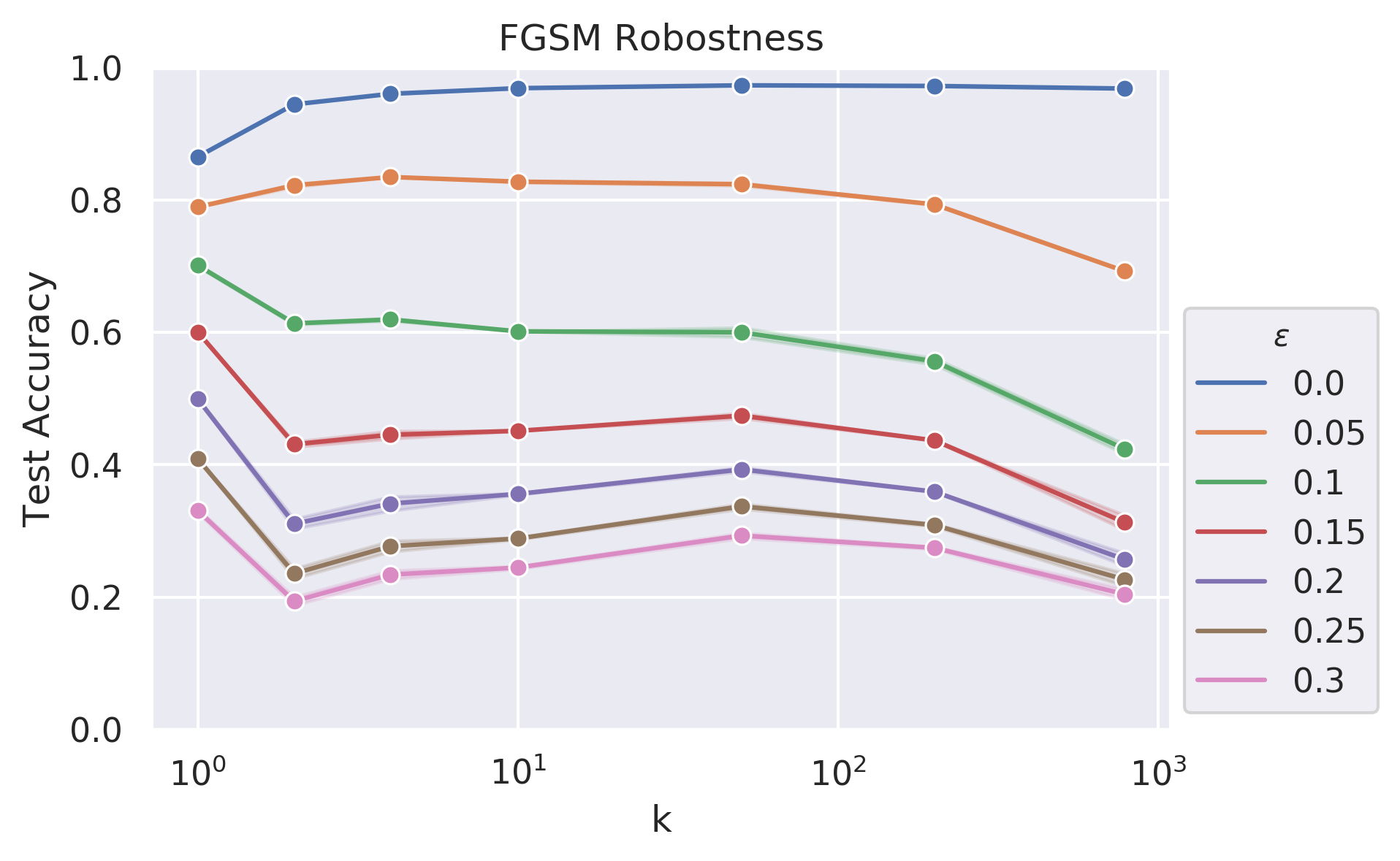}
			\includegraphics[width=0.5\linewidth]{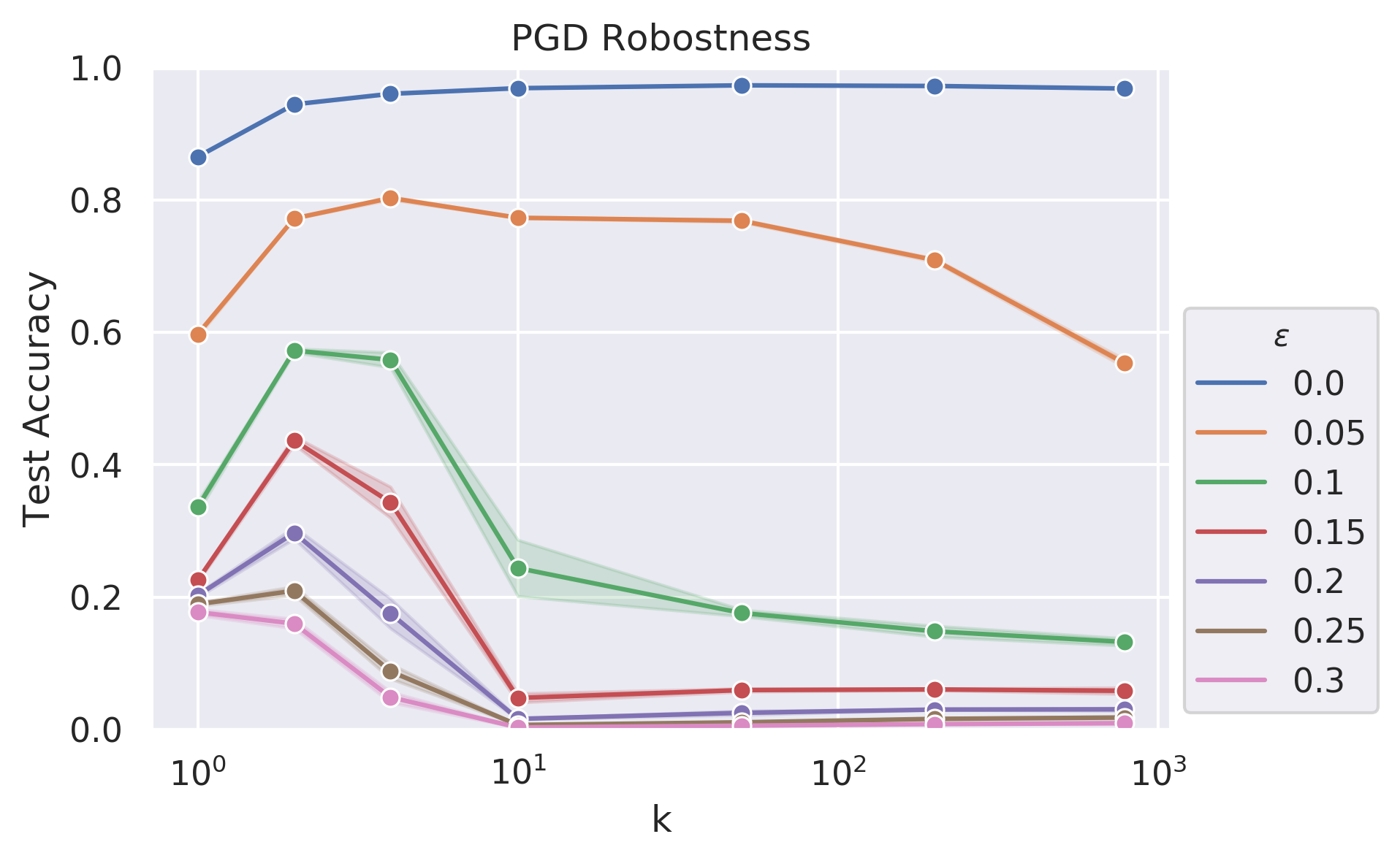}
		\end{minipage}}%
	}
\end{figure}

\paragraph{Network width}We first explore the effect of network width on adversarial robustness. The models have fully connected (FC) granule layer with different GCs ($m$ ranges from 1000 up to 200,000). We find it necessary to decrease the learning rate as the network width grows to stable the training process, therefore we heuristically set the learning rate inversely proportional to network width, thus $lr=0.5/m$. The results are shown in Fig \ref{fig:width}. We observe that the network robustness increases along with its width. Such a phenomenon is also reported by \citet{madry2017towards}, which leads to their conclusion that increasing the network capacity alone helps improve network robustness. However, as presented in the figure, our results show that the improvements are relatively subtle - and no enhancements are shown under high perturbation PGD attack. More importantly, the performance of the shallow and wide model on adversarial robustness is generally worse than the deep model. Therefore, the shallow and wide architecture reveals no advantages on adversarial robustness comparing with the deep model. The reason that such inefficient architecture appears in the cerebellum might be the cerebellum fails to evolve the sophisticated backpropagation algorithm and thus increases its learning ability by simply growing the width.

\paragraph{Long-term depression}We next study the effect of long-term depression. We repeat the experiments in network width with normalization techniques (moving average, ma) before inputs to Purkinje layers. The results are shown in Fig \ref{fig:ltd}. We do not observe meaningful improvements in robustness across all different widths.

\paragraph{Sparse connectivity}We consequently introduce sparse connectivity to the granule layer as stated above. We consider different level of sparsity $k = 1, 2, 4, 10, 50, 200, 784$. The number of granule cells is given by $m=4.8M/(k+2c)$ to control the total number of synapses, and the learning rate is decided as above. The results are shown in Fig \ref{fig:sparse}. We observe that sparse connections show limited advantages on robustness to full connections---the optimal attains when $k=2$, but in all the effect is of no significance.

\begin{figure}[htbp]
	\floatconts
	{fig:subfigex}
	{\caption{\textbf{Adversarial robustness of cerebellum on MNIST.} With the help of LTD and sparse connectivity, the cerebellum model shows improvements under low perturbations, but it is still distant from claiming the model as a robust model against adversarial examples.}\label{fig:cerebellum}}
	{%
		\subfigure[FGSM]{\label{fig:cereb_fgsm-mnist}%
			\includegraphics[width=0.5\linewidth]{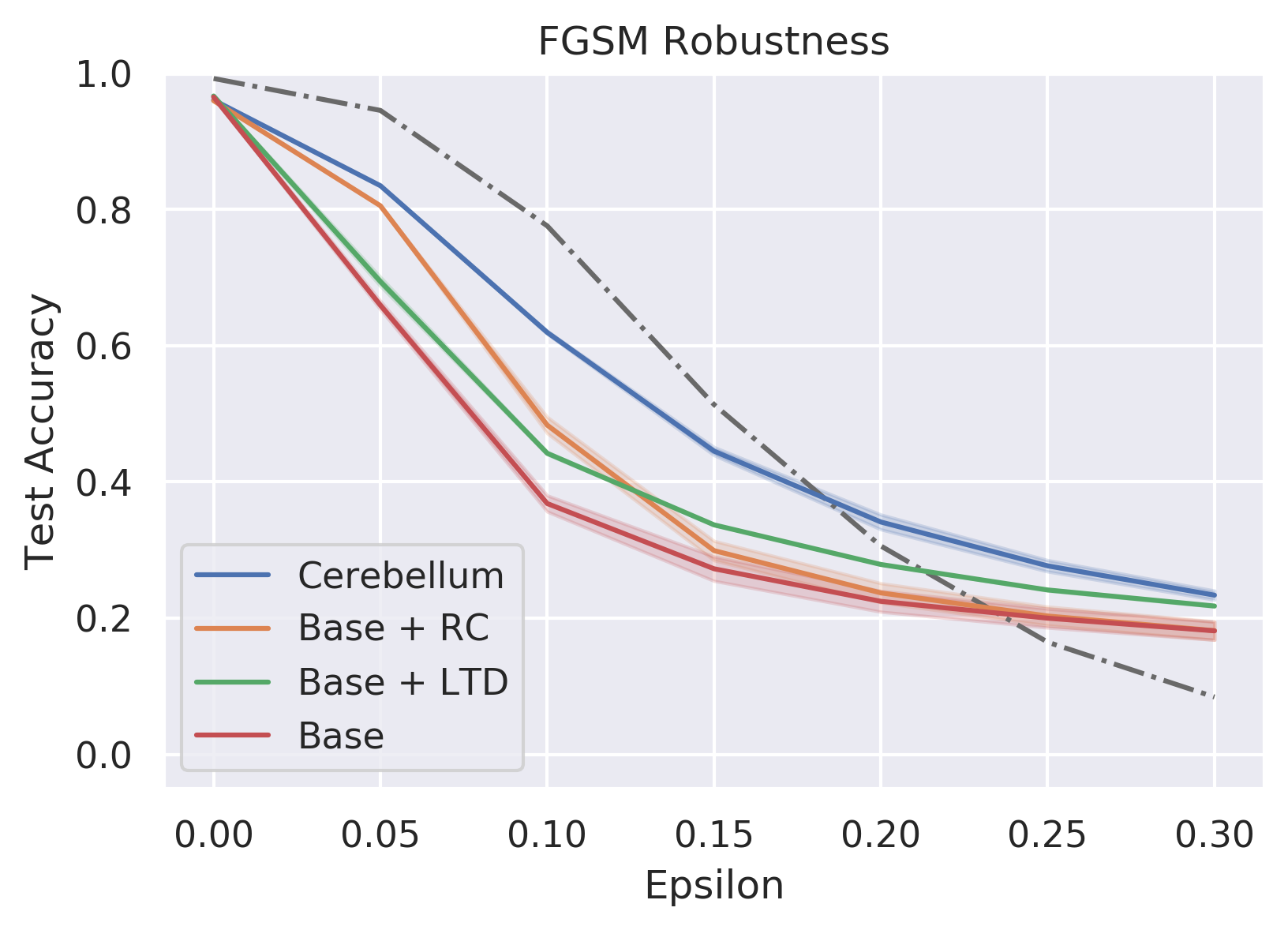}}%
		\subfigure[PGD]{\label{fig:cereb_pgd-mnist}%
			\includegraphics[width=0.5\linewidth]{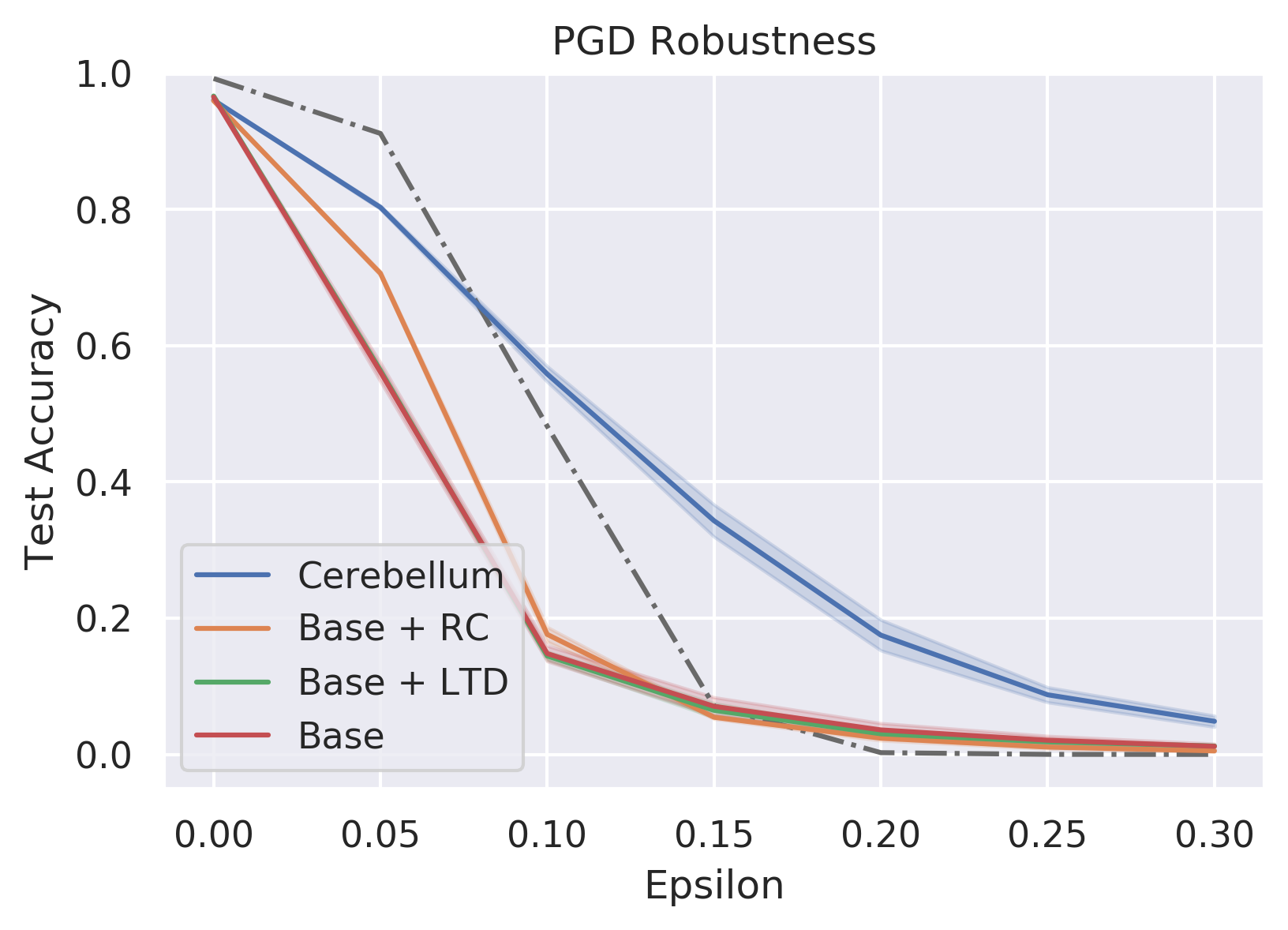}}
	}
\end{figure}

\paragraph{Cerebellum models and ablation studies on MNIST}Combining the mechanisms discussed above, we now evaluate the adversarial robustness of the cerebellum on the MNIST dataset. By with/without the long-term depression and sparse connectivity, we consider four cerebellum-like models. The base model is fully connected in the granule layer without long-term depression mechanism ($d=784,\,m=5970$), and the cerebellum model is sparse connected with long-term depression ($d=784,\,k=4,\,m=200,000$). The results are shown in Fig.~\ref{fig:cerebellum} and Table \ref{tab:mnist}. Moderate improvements in robustness are shown from the cerebellum model under low perturbations, but the model is still vulnerable to adversarial examples, especially under high perturbations.

\begin{table}[htbp]
\caption{Evaluation on the adversarial robustness of cerebellum on the MNIST dataset.}
\label{tab:mnist}
\medskip
\centering
\begin{tabular}{l|c|cccc}	
	\toprule[1.5pt]
	\multirow{2}{*}{\textbf{Model}} &  
	\multirow{2}{*}{\textbf{Test acc.}} & \multicolumn{2}{c}{\textbf{FGSM}} &
	\multicolumn{2}{c}{\textbf{PGD}} \\
	& & $\epsilon=0.1$ & $\epsilon=0.3$ & $\epsilon=0.1$ & $\epsilon=0.3$ \\
	\midrule[1pt]
	Baseline CNN & 99.2 & 77.6 & 8.4 & 48.2 & 0.0 \\
	\midrule   
	Base       & 96.5 & 36.8 & 18.1 & 14.8 & 1.2 \\
	Base + LTD & \textbf{96.7} & 44.2 & 21.8 & 14.5 & 1.2 \\
	Base + RC  & 96.0 & 48.3 & 18.2 & 17.6 & 0.5 \\
	Cerebellum & 96.1 & \textbf{62.0} & \textbf{23.4}  & \textbf{55.8} & \textbf{4.9} \\
	\bottomrule[1.5pt]
\end{tabular}
\end{table}

\paragraph{Experiments on CIFAR-10}As discussed in the proposal, the cerebellum is not suitable for directly dealing with visual stimuli. Therefore we first train a standard CNN model VGG-11 for 200 epochs as the baseline. During baseline training, we perform standard data augmenting for CIFAR-10, including random cropping and random horizontal flipping, as well as weight decay and learning rate scheduling. After acquiring a desirable visual model of CIFAR-10, we deploy the $8$ convolution layers as the visual representation extractor to train our cerebellum-like model for 10 epochs. During the cerebellum training, the same data augmenting is applied and the visual model is fixed (only serves as a representation extractor). Consequently, we perform the same attacks on the whole model. No data augmenting techniques are applied during testing and attacking. The results are summarized in Table \ref{tab:cifar10}. Similar to the MNIST dataset, no considerable improvements in robustness are found through the cerebellum-like structure.

\begin{table}[htbp]
\caption{Evaluation on the adversarial robustness of cerebellum on the CIFAR-10 dataset.}
\label{tab:cifar10}
\medskip
\centering
\begin{tabular}{l|c|cccc}	
	\toprule[1.5pt]
	\multirow{2}{*}{\textbf{Model}} & 
	\multirow{2}{*}{\textbf{Test acc.}} & \multicolumn{2}{c}{\textbf{FGSM}} &
	\multicolumn{2}{c}{\textbf{PGD}} \\
	& & $\epsilon=2/255$ & $\epsilon=8/255$ & $\epsilon=2/255$ & $\epsilon=8/255$ \\
	\midrule[1pt]
	VGG-11     & 93.0 & 54.4 & 43.1 & 35.7 & 15.4 \\
	ResNet-50  & 94.6 & 40.7 & 20.4 & 13.4 & 0.0 \\
	\midrule   
	VGG + Base       & \textbf{92.9} & 54.1 & 44.8 & 43.8 & 33.8 \\
	VGG + Base + LTD & 92.8 & \textbf{56.1} & 48.0 & 47.3 & 38.0 \\
	VGG + Base + RC  & 92.8 & 54.6 & \textbf{48.2} & \textbf{48.0} & \textbf{39.3} \\
	VGG + Cerebellum & 92.7 & 54.0 & 46.0 & 45.4 & 36.1 \\
	\bottomrule[1.5pt]
\end{tabular}
\end{table}

\section{Findings}
\label{sec:finding}
\paragraph{Shallow and wide architectures exhibit no advantage on robustness.}Opposite to the conventional hypothesis, the shallow and wide structure itself will not boost the adversarial robustness compared with the traditional deep and narrow model. The reason that the cerebellum develops such an inefficient architecture appears to be its failure in evolving the sophisticated backpropagation algorithm.

\paragraph{Cerebellum-like models are also vulnerable to adversarial attacks under batch training.}Despite the network width, LTD and sparse connectivity also do not significantly improve the robustness. In all, our experiments lead to the conclusion that the cerebellum is also susceptible to adversarial examples under batch training, unless other critical mechanisms are omitted in the classical model.

\paragraph{Intriguing adversarial robustness is discovered in the cerebellum-like model with a small batch size.}We also repeat the experiments with the training/attacking batch size equals to $1$ to mimic the biological process. Surprisingly, robustness against adversarial examples is discovered in the cerebellum-like model under this setting. Note that with the batch size equals to $1$, the experiments are $50$ times slower and shall need formidable computational resources for the PGD attack. Since the result is unexpected and beyond the original protocol, we decide to report in the appendix for future interests.

\section{Documented Modifications}
\label{sec:modification}
We have strictly followed the proposed protocol in the experimental phase. The training and attacking batch sizes are adjusted from $10$ to $50$ to accelerate the process. We also expanded the network width $m$ from $50,000$ to $200,000$ for a better evaluation of the relation. We do not include the gradient masking check since no considerable improvements have been found in robustness. No other modifications are made from the protocol.

\bibliography{yuezhang21}

\appendix
\newpage
\section{Hyperparameters}

\begin{table}[htbp]
	\caption{Hyperparameters for training and attacking the Cerebellum model.}
	\label{tab:hyper-cereb}
	\medskip
	\centering
	\begin{tabular}{p{0.25\linewidth}c}	
		\toprule[1.5pt]
		\textbf{Hyperparameter} & \textbf{Value} \\
		\midrule[1pt]  
		Train Batch      & 50 \\
		Eval Batch       & 50 \\
		Epoch            & 10 \\
		Learning rate    & $0.5/m$ \\
		Optimizer        & RMSprop \\
		Optimizer decay  & 0.99 \\
		LTD $\beta$      & 0.99 \\
		PGD steps        & 40 \\
		PGD step size    & 0.1 \\
		Seed             & \{0, 123\} \\
		\bottomrule[1.5pt]
	\end{tabular}

	\caption{Hyperparameters for pretraining the visual model on CIFAR-10.}
	\label{tab:hyper-cifar10}
	\medskip
	\centering
	\begin{tabular}{p{0.25\linewidth}c}	
		\toprule[1.5pt]
		\textbf{Hyperparameter} & \textbf{Value} \\
		\midrule[1pt]  
		Batch size       & 128 \\
		Epoch            & 200 \\
		Initial learning rate    & 0.1 \\
		Optimizer        & SGD \\
		Momentum         & 0.9 \\
		Weight decay     & 0.0001 \\
		Seed             & 0 \\
		\bottomrule[1.5pt]
	\end{tabular}
\end{table}

\section{Main Results with Proposed Batch Size}
The representative original data in Fig~\ref{fig:result-mnist} are listed in Table~\ref{tab:data-mnist}. Results are averaged over $2$ random seeds. Standard deviations are not reported as they are relatively small ($<1\%$ on accuracy) in all cases.
\begin{table}[htbp]
	\caption{Effect of network width, LTD, and sparse connectivity on MNIST (batch 50).}
	\label{tab:data-mnist}
	\medskip
	\centering
	\begin{tabular}{p{0.17\linewidth}|cc|c|cccc}
		\toprule[1.5pt]
		\multirow{2}{*}{\textbf{Model}} &
		\multirow{2}{*}{\textbf{k}} & 
		\multirow{2}{*}{\textbf{m}} & 
		\multirow{2}{*}{\textbf{Test acc.}} &
		\multicolumn{2}{c}{\textbf{FGSM}} &
		\multicolumn{2}{c}{\textbf{PGD}} \\
		& & & & $\epsilon=0.1$ & $\epsilon=0.3$ & $\epsilon=0.1$ & $\epsilon=0.3$ \\
		\midrule[1pt]
		Baseline CNN & & --- & 99.2 & 77.6 & 8.4 & 48.2 & 0.0 \\
		\midrule
		\multirow{8}{10em}{FC}       
		& &   1000   & 94.1 & 20.4 & 10.5 & 9.0 & 1.3 \\
		& &   2000   & 95.4 & 27.6 & 14.2 & 13.8 & \textbf{1.8} \\
		& &   5000   & 96.3 & 37.5 & 20.1 & 15.7 & 1.3 \\
		& &  10000   & 96.8 & 38.4 & 18.8 & \textbf{16.4} & 1.0 \\
		& &  20000   & 97.1 & 42.9 & 21.1 & 16.2 & 0.9 \\
		& &  50000   & 97.2 & 43.9 & 22.1 & 15.6 & 0.6 \\
		& & 100000   & \textbf{97.2} & \textbf{45.6} & \textbf{23.1} & 15.3 & 0.7 \\
		& & 200000   & 97.1 & 45.2 & 22.9 & 15.3 & 0.6 \\
		\midrule
		\multirow{8}{10em}{FC + LTD}       
		& &   1000   & 94.4 & 21.2 & 10.9 & 9.5 & 1.4 \\
		& &   2000   & 95.6 & 29.7 & 15.6 & 12.3 & \textbf{1.4} \\
		& &   5000   & 96.6 & 41.8 & 21.2 & 14.4 & 1.3 \\
		& &  10000   & 97.0 & 48.3 & 22.6 & 13.0 & 1.0 \\
		& &  20000   & 97.3 & 55.7 & 28.2 & 16.1 & 1.0 \\
		& &  50000   & 97.3 & 59.2 & 29.2 & 16.2 & 0.8 \\
		& & 100000   & 97.3 & 61.0 & 29.4 & 17.8 & 0.8 \\
		& & 200000   & \textbf{97.4} & \textbf{62.0} & \textbf{30.4} & \textbf{19.6} & 0.7 \\
		\midrule[1pt]
		\multirow{7}{10em}{RC}
		& 784 &   5970 & 96.5 & 33.8 & 15.9 & 13.9 & 1.3 \\   
		& 200 &  21818 & 97.1 & 43.6 & 22.3 & 15.1 & 1.0 \\
		& 50  &  68571 & \textbf{97.2} & 46.1 & 23.9 & 16.3 & 0.6 \\
		& 10  & 160000 & 96.7 & 44.2 & 19.5 & 12.1 & 0.2 \\
		& 4   & 200000 & 96.0 & 48.3 & 18.2 & 17.6 & 0.5 \\
		& 2   & 218182 & 94.2 & 54.2 & 14.9 & \textbf{24.2} & 2.0 \\
		& 1   & 228571 & 85.8 & \textbf{68.4} & \textbf{28.3} & 20.2 & \textbf{9.7} \\
		\midrule[1pt]
		\multirow{7}{10em}{RC + LTD}
		& 784 &   5970 & 96.9 & 42.4 & 20.4 & 13.2 & 0.9 \\   
		& 200 &  21818 & 97.2 & 55.6 & 27.4 & 14.8 & 0.8 \\
		& 50  &  68571 & \textbf{97.3} & 60.0 & 29.3 & 17.6 & 0.6 \\
		& 10  & 160000 & 96.9 & 60.2 & 24.5 & 24.4 & 0.3 \\
		& 4   & 200000 & 96.1 & 62.0 & 23.4 & 55.8 & 4.9 \\
		& 2   & 218182 & 94.5 & 61.4 & 19.4 & \textbf{57.3} & 16.0 \\
		& 1   & 228571 & 86.6 & \textbf{70.1} & \textbf{33.1} & 33.6 & \textbf{17.7} \\
		\bottomrule[1.5pt]
	\end{tabular}
\end{table}

\section{Intriguing Results with Small Batches}
By setting the batch size equals to $1$, we shall repeat the main experiments on MNIST. The results are summarized in Table~\ref{tab:data-mnist-1}. Surprisingly, the cerebellum-like model has exhibited considerable robustness against the PGD attack. We suspect that it is induced by not optimally setting the attacking hyperparameters for the different batch size, but are not able to perform the tuning due to the high computational demand with small batch sizes.

\begin{table}[htbp]
	\caption{Effect of network width, LTD, and sparse connectivity on MNIST (batch 1).}
	\label{tab:data-mnist-1}
	\medskip
	\centering
	\begin{tabular}{p{0.17\linewidth}|cc|c|cccc}
		\toprule[1.5pt]
		\multirow{2}{*}{\textbf{Model}} &
		\multirow{2}{*}{\textbf{k}} & 
		\multirow{2}{*}{\textbf{m}} & 
		\multirow{2}{*}{\textbf{Test acc.}} &
		\multicolumn{2}{c}{\textbf{FGSM}} &
		\multicolumn{2}{c}{\textbf{PGD}} \\
		& & & & $\epsilon=0.1$ & $\epsilon=0.3$ & $\epsilon=0.1$ & $\epsilon=0.3$ \\
		\midrule[1pt]
		Baseline CNN & & --- & 99.2 & 77.6 & 8.4 & 48.2 & 0.0 \\
		\midrule
		\multirow{8}{10em}{FC}       
		& &   1000   & 94.1 & 20.4 & 10.5 & 9.0 & 1.3 \\
		& &   2000   & 95.4 & 27.6 & 14.2 & 13.8 & \textbf{1.8} \\
		& &   5000   & 96.3 & 37.5 & 20.1 & 15.7 & 1.3 \\
		& &  10000   & 96.8 & 38.4 & 18.8 & \textbf{16.4} & 1.0 \\
		& &  20000   & 97.1 & 42.9 & 21.1 & 16.2 & 0.9 \\
		& &  50000   & 97.2 & 43.9 & 22.1 & 15.6 & 0.6 \\
		& & 100000   & \textbf{97.2} & \textbf{45.6} & \textbf{23.1} & 15.3 & 0.7 \\
		& & 200000   & 97.1 & 45.2 & 22.9 & 15.3 & 0.6 \\
		\midrule[1pt]
		\multirow{8}{10em}{FC + LTD}       
		& &   1000   & 94.4 & 21.2 & 10.9 & 9.5 & 1.4 \\
		& &   2000   & 95.6 & 29.7 & 15.6 & 12.3 & \textbf{1.4} \\
		& &   5000   & 96.6 & 41.8 & 21.2 & 14.4 & 1.3 \\
		& &  10000   & 97.0 & 48.3 & 22.6 & 13.0 & 1.0 \\
		& &  20000   & 97.3 & 55.7 & 28.2 & 16.1 & 1.0 \\
		& &  50000   & 97.3 & 59.2 & 29.2 & 16.2 & 0.8 \\
		& & 100000   & 97.3 & 61.0 & 29.4 & 17.8 & 0.8 \\
		& & 200000   & \textbf{97.4} & \textbf{62.0} & \textbf{30.4} & \textbf{19.6} & 0.7 \\
		\midrule[1pt]
		\multirow{7}{10em}{RC}
		& 784 &   5970 & 96.7 & 35.0 & 16.9 & 13.8 & 1.5 \\  
		& 200 &  21818 & 97.6 & 40.4 & 20.3 & 15.4 & 1.6 \\
		& 50  &  68571 & \textbf{97.9} & 46.9 & 21.4 & 15.2 & 0.9 \\
		& 10  & 160000 & 97.7 & 42.1 & 17.6 & 33.6 & 6.9 \\
		& 4   & 200000 & 97.0 & 48.2 & 19.2 & \textbf{33.9} & 3.3 \\
		& 2   & 218182 & 96.1 & 56.3 & 21.8 & 19.0 & 1.2 \\
		& 1   & 228571 & 83.7 & \textbf{65.4} & \textbf{26.8} & 19.3 & \textbf{9.8} \\
		\midrule[1pt]
		\multirow{7}{10em}{RC + LTD}
		& 784 &   5970 & 97.0 & 82.2 & 56.0 & 83.7 & 27.9 \\   
		& 200 &  21818 & 97.8 & 84.1 & 65.0 & 72.1 & 11.6 \\
		& 50  &  68571 & 98.0 & 86.1 & 70.0 & 75.8 & 11.4 \\
		& 10  & 160000 & 97.7 & 88.9 & \textbf{71.2} & 88.3 & 18.7 \\
		& 4   & 200000 & 97.2 & \textbf{89.0} & 67.6 & \textbf{95.2} & 48.0 \\
		& 2   & 218182 & 96.2 & 85.2 & 65.0 & 94.1 & \textbf{63.3} \\
		& 1   & 228571 & 84.9 & 69.3 & 36.1 & 55.6 & 26.7 \\
		\bottomrule[1.5pt]
	\end{tabular}
\end{table}
\end{document}